\RequirePackage{tikz}
\documentclass[sn-basic]{sn-jnl}
\usepackage[bottom]{footmisc}
\usepackage{moreverb,url}
\usepackage{subcaption}
\usepackage{multirow,array}

\pgfdeclarelayer{background}
\usepackage{bm}
\usepackage{tikz}
\usetikzlibrary{arrows.meta, shapes.geometric, calc, shadows, positioning,backgrounds,fit}

\def\shft#1{\stackon[150pt]{}{\kern -0pt #1}}
\def\shfta#1{\stackon[100pt]{}{\kern -0pt #1}}
\def\shfty#1{\stackon[100pt]{}{\kern -0pt #1}}
\def\shftya#1{\stackon[70pt]{}{\kern -0pt #1}}
\def\shftys#1{\stackon[40pt]{}{\kern -0pt #1}}
\def\shftysa#1{\stackon[10pt]{}{\kern -0pt #1}}
\tikzset{
  materia1/.style={draw, fill=fill1, text width=6.5em, text centered, minimum height=1.5em,drop shadow},
  materia2/.style={draw, fill=fill2, text width=6.5em, text centered, minimum height=1.5em,drop shadow},
  materia3/.style={draw, fill=fill3, text width=6.5em, text centered, minimum height=1.5em,drop shadow},
  etape/.style={materia1, text width=6.5em, minimum width=6.5em, minimum height=3em, rounded corners, drop shadow},
  etape2/.style={materia2, text width=6.5em, minimum width=6.5em, minimum height=3em, rounded corners, drop shadow},
  etape3/.style={materia3, text width=6.5em, minimum width=6.5em, minimum height=3em, rounded corners, drop shadow},
  linepart/.style={draw, thick, color=black!50, -LaTeX, dashed},
  line/.style={draw, thick, color=black!90, -LaTeX},
  ur/.style={draw, text centered, minimum height=0.01em},
  back group/.style={fill=yellow!20,rounded corners, draw=black!50, dashed, inner xsep=15pt, inner ysep=10pt},
  set/.style={draw,rectangle,inner sep=0pt,align=left, above=8pt},
}
\tikzset{
  groupassum/.style={draw, fill=ibmlight1, text width=6.5em, text centered, minimum height=3em},
  assum/.style={draw, fill=ibmlight13, text width=6.5em, text centered, minimum height=3em, rounded corners},
  groupcompon/.style={draw, fill=ibmlight5, text width=7em, text centered, minimum height=3em},
  compon/.style={draw, fill=ibmlight52, text width=7em, text centered, minimum height=3em,minimum width=7em, rounded corners},
  componprop/.style={draw, fill=ibmlight52,minimum width=9.5em, text width=11em, text centered, text
    depth  = 2cm, minimum height=8em, rounded corners},
  componprop2/.style={draw, fill=ibmlight52,minimum width=9.5em, text centered, text
    depth  = 1.5cm, minimum height=3.5em, rounded corners},
  property/.style={draw, fill=ibmlight53, text width=8em, text centered, minimum height=2em},
  grouptask/.style={draw, fill=ibmlight2, text width=7em, text centered, minimum height=3em},  
  task/.style={draw, fill=ibmlight22, text width=7em, text centered, minimum height=3em, rounded corners},
  methods/.style={draw, fill=ibmlight23, text width=7em, text centered, minimum height=3em, rounded corners},  
  ass/.style={assum, text width=6.5em, minimum width=6em, minimum height=3em},
datassum/.style={draw, fill=ibmlight1, text width=7em, text centered, minimum height=3em, rounded corners},
dgpassum/.style={draw, fill=ibmlight12, text width=7em, text centered, minimum height=3em, rounded corners}, 
causalassum/.style={draw, fill=ibmlight14, text width=7em, text centered, minimum height=3em, rounded corners}
}

\newcommand{\assum}[2]{node (p#1) [assum] {#2}}
\newcommand{\compon}[2]{node (p#1) [compon] {#2}}
\newcommand{\task}[2]{node (p#1) [task] {#2}}
\usetikzlibrary{arrows}
\newcommand{\mycomment}[1]{}

\definecolor{ibm1}{HTML}{648FFF}
\definecolor{ibm2}{HTML}{785EF0}
\definecolor{ibm3}{HTML}{DC267F}
\definecolor{ibm4}{HTML}{FE6100}
\definecolor{ibm5}{HTML}{FFB000}

\definecolor{fill1}{HTML}{CFD5E8}
\definecolor{stroke1}{HTML}{324068}
\definecolor{fill2}{HTML}{F3F6FB}
\definecolor{stroke2}{HTML}{8594B6}
\definecolor{fill3}{HTML}{E5E5E5}
\definecolor{stroke3}{HTML}{4A4849}
\definecolor{line}{HTML}{818689}

\definecolor{ibmlight1}{HTML}{F4F7FE}
\definecolor{ibmlight12}{HTML}{E2EAFD}
\definecolor{ibmlight13}{HTML}{D7E0F9}
\definecolor{ibmlight14}{HTML}{C0CFF8}
\definecolor{ibmlight2}{HTML}{F7F5FB}
\definecolor{ibmlight22}{HTML}{EBE8F7}
\definecolor{ibmlight23}{HTML}{e0dcf3}

\definecolor{ibmlight5}{HTML}{f2e8ce}
\definecolor{ibmlight52}{HTML}{fdf3db}
\definecolor{ibmlight53}{HTML}{FCEBC3}

\definecolor{ibmlight5}{HTML}{f2e8ce}
\definecolor{ibmlight54}{HTML}{fce5cd}
\definecolor{ibmlight55}{HTML}{f7d9d9}

\definecolor{ibmlight3}{HTML}{f7e9f0}
\definecolor{ibmlight4}{HTML}{f7ebe2}

\jyear{2024}%

\DeclareMathOperator*{\argmax}{arg\,max}

\theoremstyle{thmstyleone}%
\theoremstyle{thmstyletwo}%
\newtheorem{example}{Example}%

\theoremstyle{thmstylethree}%
\newtheorem{definition}{Definition}%

\begin{document}

\title[Graphical Models for Decision-Making: Integrating Causality and Game Theory]{Graphical Models for Decision-Making: Integrating Causality and Game Theory}

\author[1,2]{\fnm{Maarten} C. \sur{Vonk}}\email{maartencvonk@gmail.com}
\author[3]{\fnm{Mauricio} \sur{Gonzalez Soto}}
\author[1]{\fnm{Anna V.} \sur{Kononova}}

\affil[1]{\orgdiv{LIACS}, \orgname{Leiden University}, \orgaddress{\street{Einsteinweg 55}, \city{Leiden}, \postcode{2333CC} \country{Netherlands}}}

\affil[2]{\orgname{The Hague Centre of Strategic Studies}, \orgaddress{\street{Lange Voorhout 1}, \city{The Hague}, \postcode{2514EA}, \country{Netherlands}}}

\affil[3]{\orgname{Tecnologico de Monterrey, Escuela de Ingenieria y Ciencias}, \orgaddress{\street{Eugenio Garza Sada 2501 Sur, Col. Tecnologico},\city{Monterrey, N.L.},\postcode{64849},\country{Mexico}}}

\abstract{Causality and game theory are two influential fields that contribute significantly to decision-making in various domains. Causality defines and models causal relationships in complex policy problems, while game theory provides insights into strategic interactions among stakeholders with competing interests. Integrating these frameworks has led to significant theoretical advancements with the potential to improve decision-making processes. However, practical applications of these developments remain underexplored. To support efforts toward implementation, this paper clarifies key concepts in game theory and causality that are essential to their intersection, particularly within the context of probabilistic graphical models. By rigorously examining these concepts and illustrating them with intuitive, consistent examples, we clarify the required inputs for implementing these models, provide practitioners with insights into their application and selection across different scenarios, and reference existing research that supports their implementation. We hope this work encourages broader adoption of these models in real-world scenarios.}


\maketitle

\section{Introduction}



Causality involves defining and modeling causal relationships between variables, aiming to determine how changes in one variable influence another. The key distinction between causal and associative models is that causal models focus on intervention rather than mere observation \citep{Spirtes2000,tesisgonzalezsoto}. Once a causal relationship between two events is established, actively manipulating the cause enables more reliable outcome predictions compared to relying solely on correlations between observations. Developing a causal model of an environment and leveraging it for decision-making enables explanations beyond what an associative model can provide—it allows one to ask why. Once learned, a causal model is independent of the agent and its preferences, making the acquired knowledge transferable to similar domains where understanding causal relationships proves beneficial \citep{pearl2018book,gonzalez2023learning}. Because of this practical significance and broad applicability, causal analysis has permeated diverse research fields, including medical treatment \citep{Shalit2020}, policy-making \citep{Kreif2019}, social sciences \citep{Sobel2014}, epidemiology \citep{Halloran1995}, and cybersecurity \citep{Andrew2022, Dhir2021}.

On the other hand, game theory studies strategic interactions between rational decision-makers. It provides a framework to formalize the behavior resulting from interactions between competitive or cooperative agents. As a result, it has percolated the field of international security \citep{LINDELAUF2013230}, cyber security \citep{do2017game}, biology \citep{leimar2023game}, and climate change \citep{wood2011climate}.

Given the strengths of both fields, recent research has integrated them, allowing for causal interventions in settings where strategic interactions play a critical role \citep{hammond2023reasoning,soto2019choosing}.  Despite significant theoretical advances in causal game theory, practical applications remain limited.

Several studies indicate that researchers often avoid employing causal and game-theoretic models due to difficulties in both comprehension and implementation \citep{busetti2023causality, musci2019ensuring, hermans2014usefulness}. This challenge is even more pronounced when integrating causal reasoning with game-theoretic modeling, as the absence of established frameworks complicates their combined application. Consequently, there have been increasing calls to bridge the gap between theory and practice through structured dialogues and closer collaborations between practitioners and methodologists \citep{shubik2011present, musci2019ensuring}.

To address this gap, this paper provides a structured framework that clarifies key concepts in causality, game theory, and their intersection. By consistently applying a single practical example and referencing relevant research for implementation, we offer a concrete guide to navigating the complexities of integrating causal reasoning with game-theoretic modeling. This approach aims to bridge the divide between theory and practice, equipping practitioners with clearer implementation strategies and fostering closer collaboration between researchers and methodologists.

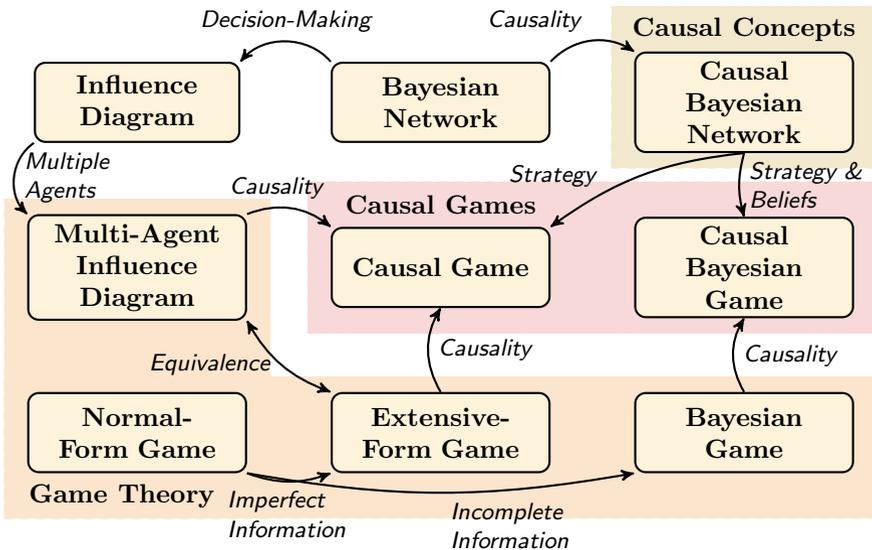
\begin{figure}[!b]
\centering
\tikzset{methods/.style={draw, fill=ibmlight23, text width=7em, text centered, minimum height=3em, rounded corners}}
\def\blockdist{2.3}
\def\edgedist{2.5}
\def\blockdist{2.3}
\def\edgedist{2.5}
\begin{tikzpicture}[->,>=stealth',auto,node distance=3cm,
  thick,main node/.style={circle,draw,font=\sffamily\Large\bfseries}]

    \node (id) [compon] {\textbf{Influence Diagram}};
    \path (id.center)+(4,0) node (bn) [compon,text width=7.5em] {\textbf{Bayesian Network}};
    \path (id.center)+(8,0) node (cbn) [compon,text width=7.5em] {\textbf{Causal Bayesian Network}}; 
    \path (id.center)+(0,-2.2) node (maid) [compon,text width=7.5em ] {\textbf{Multi-Agent Influence Diagram}};
    \path (id.center)+(4,-2.2) node (cg) [compon, text width=7.5em] {\textbf{Causal Game}};
    \path (id.center)+(8,-2.2) node (cbg) [compon, text width=7.5em] {\textbf{Causal Bayesian Game}};
    \path (id.center)+(0,-4.4) node (nmf) [compon,text width=7.5em ] {\textbf{Normal-Form Game}};
    \path (id.center)+(4,-4.4) node (efg) [compon, text width=7.5em] {\textbf{Extensive-Form Game}};
    \path (id.center)+(8,-4.4) node (bg) [compon, text width=7.5em] {\textbf{Bayesian Game}};
    \path (cbn.north)+(0,+0.3) node (assum) {\textbf{Causal Concepts}};
    \path (cg.north)+(0,+0.3) node (assum) {\textbf{Causal Games}};
    \path (efg.south)+(-4.2,-0.3) node (assum) {\textbf{Game Theory}};

    \path[every node/.style={font=\sffamily\small\itshape}]
        (bn.north west) edge[bend right=50] node [above] {Decision-Making} (id.north east);    
    \path[every node/.style={font=\sffamily\small\itshape}]
        (id.south west) edge[bend right=50] node [right,text width=2em] {Multiple Agents} (maid.north west);
    \path[every node/.style={font=\sffamily\small\itshape}]
        (bn.north east) edge[bend left=40] node [above left] {Causality} (cbn.north west);
    \path[every node/.style={font=\sffamily\small\itshape}]
        (nmf.south east) edge[bend right=20] node [below,text width=4.6em] {Imperfect Information} (efg.south west);
    \path[every node/.style={font=\sffamily\small\itshape}]
        (nmf.south east) edge[bend right=10] node [below right,text width=4.6em] {Incomplete Information} (bg.south west);
    \path[every node/.style={font=\sffamily\small\itshape}]
        (maid.north east) edge[bend left=30] node [above,text width=4.6em] {Causality} (cg.north west);
    \path[<->] (maid.south east) edge[bend right=20] 
        node[above, font=\sffamily\small\itshape, left, text width=4.6em] {Equivalence} 
        (efg.north west);
    \path[every node/.style={font=\sffamily\small\itshape}]
        (efg.north) edge[bend left=30] node [right,text width=4.6em] {Causality} (cg.south);
    \path[every node/.style={font=\sffamily\small\itshape}]
        (bg.north) edge[bend left=30] node [right,text width=4.6em] {Causality} (cbg.south);
    \path[every node/.style={font=\sffamily\small\itshape}]
        (cbn.south) edge[bend right=15] node [left,text width=4.6em] {Strategy} (cg.north east);
    \path[every node/.style={font=\sffamily\small\itshape}]
        (cbn.south) edge[bend right=15] node [right,text width=6em] {Strategy \& Beliefs} (cbg.north);
    \begin{pgfonlayer}{background}
        \path (maid.north west)+(-0.3,0.2) node (a) {};
        \path (nmf.south east)+(+0.3,-0.6) node (b) {};
        \path[fill=ibmlight54, draw=ibmlight54, dashed]
            (a) rectangle (b);
            
        \path (cbn.north west)+(-0.3,0.6) node (a) {};
        \path (cbn.south east)+(0.3,-0.2) node (b) {};
        \path[fill=ibmlight5, draw=ibmlight5, dashed]
            (a) rectangle (b);

        \path (cg.north west)+(-0.3,0.6) node (a) {};
        \path (cbg.south east)+(0.3,-0.2) node (b) {};
        \path[fill=ibmlight55, draw=ibmlight55, dashed]
            (a) rectangle (b);

        \path (nmf.north west)+(-0.3,0.2) node (a) {};
        \path (bg.south east)+(0.3,-0.6) node (b) {};
        \path[fill=ibmlight54, draw=ibmlight54, dashed]
            (a) rectangle (b);
    \end{pgfonlayer}
\end{tikzpicture}
\caption{Scope of this paper: the yellow blocks represent key concepts discussed within each domain: causality, game theory, and causal game theory. The concepts are grouped into categories that highlight their primary associations, indicated with background in different colors. Arrows indicate the possible extension or adaptation that allows for the transition from one concept to another.}
\label{fig:scope}
\end{figure}

More specifically, we review the connection between causality and game theory in the context of \emph{probabilistic graphical models} (PGMs) \citep{Koller2009,sucar2020probabilistic}. We focus on PGMs as they provide a structured approach to bridging the gap between theoretical advancements and practical implementation, particularly due to their intuitive representation of dependencies \citep{pernkopf2014introduction} and their capacity to unify diverse cognitive processes through a shared representational framework \citep{danks2014unifying}. This paper uniquely unifies the connections between causal and game-theoretical concepts within a single mathematical framework, which ensures conceptual clarity and enhances interdisciplinary interoperability. Through this detailed examination of their mathematical specifics, supported by illustrative examples, we provide a structured framework that bridges theoretical understanding and practical implementation. While exploring the mathematical details of various game-theoretic models, causal models, and causal games might seem counterintuitive for practical purposes, it is necessary to articulate the distinctions between these models more precisely. This distinction not only helps practitioners select the most suitable model for their specific application but also clarifies the specific information required to work with these models effectively. Without delving into specifics, we point to research that discusses techniques for eliciting the required information. Finally, we discuss further considerations and insights to help surmount practical implementation challenges. The conceptual scope of this paper is further illustrated in Figure \ref{fig:scope}, which categorizes the key concepts discussed across causality, game theory, and causal game theory.

Besides a practical guide to causal discovery \citep{Malinsky2018}, there have been papers written about the construction of influence diagrams along with utilities \citep{BIELZA2010354} and about representation restrictions imposed by influence diagrams \citep{bielza2009modeling}. To the best of our knowledge, our work is the first to address the causal game-theoretical models from a practical point of view. 

Although doubts concerning the assumption of agent rationality \citep{sen1977rational, GINTIS2000311} have led to skepticism regarding the applicability of game-theoretic models \citep{vonNeumannMorgenstern2004, rubinstein2012, rapoport1962experimental}, our paper refrains from entering this philosophical debate. 

The structure of this paper is as follows. Section \ref{sec:got} presents game-theoretic models, their solution concepts, and illustrative applications, along with a practical guide to their implementation. Section \ref{sec:caus} explores causality, covering topics from causal discovery to causal inference and their practical applications. The intersection of causality and game theory is examined in Section \ref{sec:int}. Finally, Section \ref{sec:conc} reflects on the key insights and outlines directions for future research.


\section{Game Theory} \label{sec:got}
As the aim of this paper is to connect causality with game theory, we \textit{restrict} ourselves to the game-theoretical components that have extensions in causality. Therefore, we reflect on three different types of games: the normal-form game, the extensive-form game, and the Bayesian game. A normal-form game models strategic interactions in which players select actions simultaneously without knowledge of other players' choices, making it well-suited for competitive scenarios such as pricing strategies. An extensive-form game represents sequential decision-making, where players act in a structured order, as in negotiations. A Bayesian game incorporates uncertainty, allowing players to make decisions based on private beliefs about unknown factors, making it particularly applicable to auctions.

After introducing preliminaries on game theory, we continue by outlining formal game definitions and relevant solution concepts. Furthermore, we explore the applicability of these game forms by analyzing similar examples across different scenarios and outline practical steps for modeling with these forms. Finally, we address the challenges associated with each form and discuss their practical utility. An overview of the game forms discussed, their concomitant characteristics, and required information for model implementation are summarized in Figure \ref{fig:gtprac}.

\subsection{Preliminaries} 
A game models strategic interactions among multiple decision-makers, where each decision-maker selects actions to maximize their own objectives while considering the choices of others. It provides a structured framework to analyze decision-making in competitive or cooperative settings. The set of rational decision-makers is called the \emph{agents} or \emph{players}. A game will be denoted by $\Gamma$ and usually consists of \emph{players} $M$, \emph{action profiles} \(\bm{A} = \{A^1, A^2, \dots , A^m\}\), \emph{utility functions} \(\bm{U} = \{u^1, u^2, \dots, u^m\}\), and a graph $G=(\bm{V},\bm{E})$. Furthermore, a \textit{(behavioral) strategy} for a player in a game is a complete plan of action that specifies the move or decision the player will make in every possible situation they might face within the game. Formally, a strategy for player \( i \in M \) is defined as a function 
\( \sigma^i: A^i \to \Delta(A^i) \), where \( \Delta(A^i) \) denotes the set of probability 
distributions over the set of available actions \( A^i \). This means that \( \sigma^i \in  \bm{\Sigma}^i \) assigns 
to each action in \( A^i \) a probability such that the probabilities sum to 1, representing the likelihood of the player choosing each action. A strategy is \textit{pure} when each \(\sigma^{i}(a) \in \{0, 1\}\) and \textit{fully stochastic} when \(\sigma^{i}(a) > 0\), for all \(a \in A^i\). A \textit{strategy profile} \(\sigma = (\sigma^1, \ldots, \sigma^m)\) is a tuple of strategies, and \(\sigma^{-i} = (\sigma^1, \ldots, \sigma^{i-1}, \sigma^{i+1}, \ldots, \sigma^m)\) denotes the \emph{partial strategy} profile of all agents other than \(i\), hence \(\sigma = (\sigma^i, \sigma^{-i})\).

\subsection{Normal-Form Game}
First, we write down the definition of a normal-form game and its associated solution concept. Then we provide an example.
\begin{definition}
A \textit{normal-form game} (NFG) is a tuple \(\Gamma = (M, \bm{A}, \bm{U})\) for which:
\begin{itemize}
    \item \(M = \{1, \ldots, m\}\) is a set of agents.
    \item \(\bm{A} = \{A^1, A^2, \dots , A^m\}\) is the set of action profiles, where \(A^i\) denotes the set of actions available to agent \(i \in M\).     
    \item \(\bm{U} = \{u^1, u^2, \dots, u^m\}\) is a set of utility functions where \(u^i: \bm{A} \rightarrow \mathbb{R}\) is the payoff function for agent \(i \in M\), representing the payoff that agent \(i\) receives.

\end{itemize}
\end{definition}

\begin{definition}
A strategy profile\footnote{In normal-form games, behavioral strategies are known as \emph{mixed strategies}.} \(\hat{\sigma} = (\hat{\sigma}^1, \ldots, \hat{\sigma}^n)\) is a \textit{Nash equilibrium} if for every player $i\in \{1,\ldots,m \}$:
\begin{align*}
    \hat{\sigma}^i\in \argmax_{{\sigma}^i\in \bm{\Sigma}^i}u^i(\sigma^i,\hat{\sigma}^{-i}).
\end{align*}

\end{definition}

\begin{figure}[!b]
    \centering
    \begin{minipage}[b]{.35\textwidth}
        \centering
        \begin{subfigure}[b]{\textwidth}
            \centering
            \begin{tikzpicture}[
    bluebox/.style={draw, rectangle, fill=blue!30, minimum size=7mm, node distance=1.5cm},
    redbox/.style={draw, rectangle, fill=red!30, minimum size=7mm, node distance=1.5cm},
    bluediamond/.style={draw, diamond, fill=blue!30, minimum size=7mm, node distance=1.5cm},
    reddiamond/.style={draw, diamond, fill=red!30, minimum size=7mm, node distance=1.5cm}
]

\node[redbox] (a1) {$A^1$};
\node[bluebox, right of=a1] (a2) {$A^2$};

\node[reddiamond, below of=a1] (u1) {$u^1$};  
\node[bluediamond, below of=a2] (u2) {$u^2$}; 

\draw[->] (a1) to (u2);
\draw[->] (a2) to (u1);
\draw[->] (a1) to (u1);
\draw[->] (a2) to (u2);

\end{tikzpicture}
            \subcaption{normal-form game}
            \label{fig:normal-form}
        \end{subfigure}\\
    \end{minipage}%
    \begin{minipage}[b]{.35\textwidth}
        \centering
        \begin{subfigure}[b]{\textwidth}
            \centering
            \begin{tikzpicture}[
    redbox/.style={draw, rectangle, fill=red!30, minimum size=7mm, node distance=1.5cm},
    bluebox/.style={draw, rectangle, fill=blue!30, minimum size=7mm, node distance=1.5cm},
    reddiamond/.style={draw, diamond, fill=red!30, minimum size=7mm, node distance=1.5cm},
    bluediamond/.style={draw, diamond, fill=blue!30, minimum size=7mm, node distance=1.5cm}
]

\node[redbox] (a1) {$A^1$};
\node[reddiamond, below of=a1] (u1) {$u^1$};  
\node[bluebox, right of=a1] (a2) {$A^2$};
\node[bluediamond, below of=a2] (u2) {$u^2$}; 

\draw[->] (a1) to (a2);
\draw[->] (a1) to (u2);
\draw[->] (a2) to (u1);
\draw[->] (a1) to (u1);
\draw[->] (a2) to (u2);

\end{tikzpicture}
            \subcaption{extensive-form game}
            \label{fig:extensive-form}
        \end{subfigure}\\
    \end{minipage}%
    \begin{minipage}[b]{.30\textwidth}
        \centering
        \begin{subfigure}[b]{\textwidth}
            \centering            
            \begin{tikzpicture}[
    redbox/.style={draw, rectangle, fill=red!30, minimum size=7mm, node distance=1.5cm},
    bluebox/.style={draw, rectangle, fill=blue!30, minimum size=7mm, node distance=1.5cm},
    reddiamond/.style={draw, diamond, fill=red!30, minimum size=7mm, node distance=1.5cm},
    bluediamond/.style={draw, diamond, fill=blue!30, minimum size=7mm, node distance=1.5cm},
    circlenode/.style={draw, circle, fill=white, minimum size=7mm, node distance=1.5cm}
]

\node[redbox] (a1) {$A^1$};
\node[circlenode, shift={(-0.3cm,0cm)}, above right of=a1] (t) {$T$}; 
\node[reddiamond, below of=a1] (u1) {$u^1$};  
\node[bluebox, right of=a1] (a2) {$A^2$};
\node[bluediamond, below of=a2] (u2) {$u^2$}; 

\draw[->] (a1) to (u2.west);
\draw[->] (a2) to (u1.east);
\draw[->] (a1) to (u1);
\draw[->] (a2) to (u2);
\draw[->] (t) to (a1.north);
\draw[->] (t) to (a2.north);
\draw[->] (t) to (u1.north east);
\draw[->] (t) to (u2.north west);

\end{tikzpicture}
            \subcaption{Bayesian game}
            \label{fig:bayesian-form}
        \end{subfigure}\\
    \end{minipage}%
    \caption{The relations of the normal-form game (a), extensive-form game (b), and Bayesian game (c). These relations correspond to Example \ref{ex:normal-form} for (a), Examples \ref{ex:extensive-form} and \ref{ex:efgsubgame} for (b), and Example \ref{ex:bayesian-form} for (c). In the normal-form game, the deterring and attacking agents make independent decisions that determine their utilities. In contrast, the attacker’s decisions are shaped by the deterrer's actions in the extensive-form game. Lastly, in the Bayesian game, the agents' decisions and utilities are influenced by their individual types and their beliefs about the opponent's type.}
    \label{fig:game_theory-examples}
\end{figure}
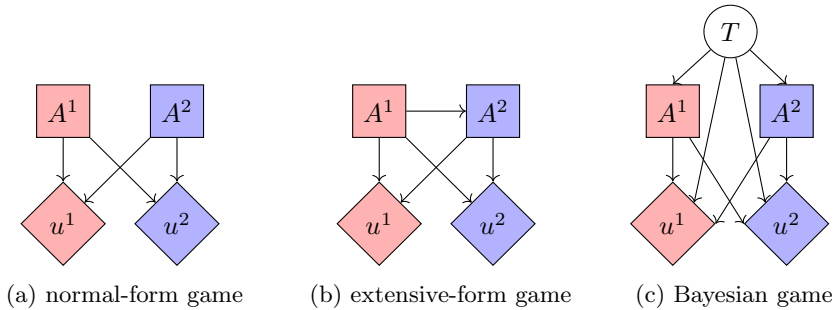

\begin{example}[Deterring Game in Normal-Form] \label{ex:normal-form}
Suppose a game models a strategic interaction between a deterring agent and its attacking adversary. The deterring agent threatens retaliation if attacked, but whether the agent is willing to follow through ($d$) or is bluffing ($\neg d$) is determined by action $A^1$. Simultaneously, the adversary decides whether to attack ($a$) or not ($\neg a$), denoted by action $A^2$. The game can be illustrated by Figure \ref{fig:normal-form} and the game matrix is displayed in Table \ref{tab:nato_nfg}. The two pure Nash equilibria are ($\neg a$, $d$) and ($a$, $\neg d$).     
\end{example}

\begin{table}
\centering
\caption{Utilities in Deterring Game (or Game of Chicken) in Normal-Form.}
\begin{tabular}[c]{cc|c|c|}
  & \multicolumn{1}{c}{} & \multicolumn{2}{c}{Adversary} \\
  & \multicolumn{1}{c}{} & \multicolumn{1}{c}{$a$}  & \multicolumn{1}{c}{$\neg a $} \\
\cline{3-4}
  \multirow{2}*{Deterrer}  & $d$ & $(-1000, -1000)$ & $(1, -1)$ \\
\cline{3-4}
  & $\neg d$ & $(-1, 1)$ & $(0, 0)$ \\
\cline{3-4}
\end{tabular}
\label{tab:nato_nfg}
\end{table}

Although the deterring agent acts first by issuing a threat of retaliation, the game-theoretical model does not consider the act of threatening as a strategic decision. Instead, the decision of interest is whether the deterring agent follows through on the threat. Since this decision does not occur after the attacker's move, both actions are effectively simultaneous and can be represented using a normal-form game. Nonetheless, it may very well be that acts do happen sequentially. In this case, a more refined game form is necessary, which is the extensive-form game. 

\subsection{Extensive-Form Game}

First, we write down the definition of an extensive-form game, followed by an example and the relevance of a solution concept known as a subgame perfect equilibrium. We follow the definition of \cite{hammond2023reasoning}.

\begin{definition}
An \textit{extensive-form game} (EFG) is a tuple \(\Gamma = (M, G, \bm{P}, \bm{A}, \lambda, \bm{I}, \bm{U})\) for which:
    \begin{itemize}
    \item \(M = \{1, \ldots, m\}\) is a set of agents.
    \item \(G = (\bm{V}, \bm{E})\) is a rooted tree, where the nodes \(\bm{V}\) are partitioned into sets \(\bm{V^0},\bm{V^1}, \ldots, \bm{V^n}, \bm{T}\). In this case, \(\bm{T}\) are the leaves or terminal nodes of \(G\), \(\bm{V^0}\) are chance nodes, and \(\bm{V^i}\) are the decision nodes controlled by agent \(i \in M\). The nodes are connected by edges \(\bm{E}\).
    \item $\bm{P} = \{P_1, P_2, \ldots, P_{\lvert \bm{V}^0 \rvert}\}$ represents a set of probability distributions $P_j$ defined over the children of each chance node $V^0_j$, denoted as $\textbf{\text{ch}}(V^0_j)$, for $j = 1, 2, \ldots, \lvert \bm{V}^0 \rvert$.
    \item $\bm{A}$ represents the set of action profiles, where $A^i_j \subseteq \bm{A}$ indicates the set of actions available at $V^i_j \in \bm{V^i}$.
    \item $\lambda : \bm{E} \to \bm{A}$ is a labelling function that assigns each edge $(V^i_j, V^k_l)$ to an action $a \in A^i_j$.
    \item $\bm{I} = \{\bm{I}^1, \ldots, \bm{I}^m\}$ represents a collection of information sets, which partition the decision nodes controlled by agent $i$. Each information set $I^i_j \in \bm{I}^i$ is defined such that, for all $V^i_k, V^i_l \in I^i_j$, the available actions at these nodes are identical, i.e., $A^i_k = A^i_l$.
    \item \(\bm{U} = \{u^1, u^2, \dots, u^m\}\) is a set of utility functions where \(u^i: \bm{T} \rightarrow \mathbb{R}\) is the payoff function for agent \(i \in M\), representing the payoff that agent \(i\) receives.
\end{itemize}
\end{definition}

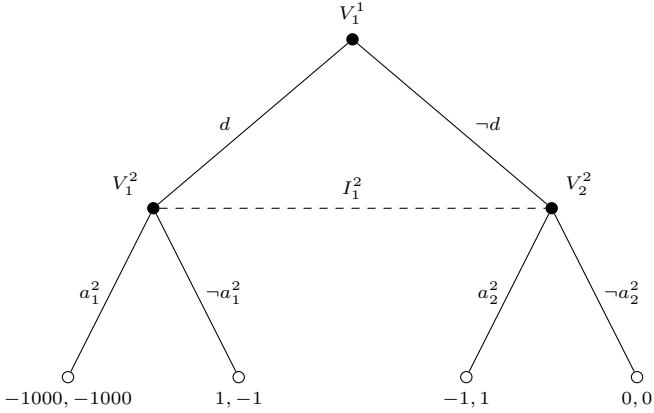
\begin{figure}[!t]
    \centering
    \begin{tikzpicture}[scale=1.5,font=\footnotesize]
\tikzstyle{solid node}=[circle,draw,inner sep=1.5,fill=black]
\tikzstyle{hollow node}=[circle,draw,inner sep=1.5]
\tikzstyle{level 1}=[level distance=15mm,sibling distance=3.5cm]
\tikzstyle{level 2}=[level distance=15mm,sibling distance=1.5cm]
\tikzstyle{level 3}=[level distance=15mm,sibling distance=1cm]

\node(0)[solid node,label=above:{$V_1^1$}]{} 
    child{node[solid node,label=above left:{$V_1^2$}]{}
        child{node[hollow node,label=below:{$-1000, -1000$}]{} edge from parent node[left] {$a_1^2$}}
        child{node[hollow node,label=below:{$1, -1$}]{} edge from parent node[right] {$\neg a_1^2$}}
        edge from parent node[left,xshift=-5] {$d$}
    }
    child{node[solid node,label=above right:{$V_2^2$}]{}
        child{node[hollow node,label=below:{$-1, 1$}]{} edge from parent node[left] {$a_2^2$}}
        child{node[hollow node,label=below:{$0, 0$}]{} edge from parent node[right] {$\neg a_2^2$}}
        edge from parent node[right,xshift=5] {$\neg d$}
    }
;
\draw[dashed] (0-1) -- (0-2) node[midway, above] {$I^2_1$};
\end{tikzpicture}
    \label{mixedc}
    \caption{Deterring Game in Extensive-Form}
    \label{fig:nato_game}
\end{figure}

\begin{example}[Deterring Game in Extensive-Form] \label{ex:extensive-form}
Suppose a game models a strategic interaction between a deterring agent and its attacking adversary. This time, the deterring agent aims to deter by threatening retaliation denoted by action $A^1$ in node $V_1^1$, which it is willing to follow through ($d$) or is bluffing ($\neg d$). Subsequentially, the adversary then acts $A^2$ to decide whether to attack ($a$) or not ($\neg a$) in nodes $(V_1^2, V_2^2)$. Since the adversary does not have knowledge about the credibility of the deterrence effort, both nodes $(V_1^2, V_2^2)$ are in the same information set, $I_1^2$. The dependencies of the game can be illustrated by Figure \ref{fig:extensive-form} and the game tree of Figure \ref{fig:nato_game}. 
\end{example}

Should the adversary have full information about the action of the deterring agent, a new game arises in which the adversary evaluates its actions based on this knowledge. Formally, this is known as a \emph{subgame}.

\begin{definition}
    A \textit{subgame} of an EFG is a game with a game tree $G'=(\bm{V}',\bm{E}')$ that is restricted to a node and its descendants such that any information set is either completely in the subgame or completely out the subgame.
\end{definition}

\begin{definition} \label{def:spe}
    A \emph{subgame perfect equilibrium} is a strategy profile $\sigma$ such that for every subgame, $\sigma$ constitutes a Nash equilibrium of the subgame.
\end{definition}

Subgame perfect equilibria are relevant for the exclusion of non-credible threats, which are threats that a rational player has no incentive to carry out in later stages of the game. This is illustrated by the following example:

\begin{example}[Deterring Subgame in Extensive-Form] \label{ex:efgsubgame}
Consider a slightly modified version of Example \ref{ex:extensive-form} under conditions of perfect information, where the adversary recognizes the deterrer's commitment to enforce the threat, as shown in Figure \ref{fig:extensive-form}. The Nash equilibria are $((\neg a_1^2, \neg a_2^2),d), ((\neg a_1^2, a_2^2),d)$ and $((a_1^2,a_2^2),\neg d)$. In the subgame, the deterring agent anticipates that the adversary's threat to attack is not credible in the case of committed punishment, leaving $((\neg a_1^2, a_2^2),d)$ as the only subgame perfect equilibrium.
\end{example}

While extensive-form games can deal with \emph{imperfect information}, which is uncertainty about the history of the game, they cannot account for \emph{incomplete information}, which comprises uncertainty about the structure of the game. This is where Bayesian games come in.

\subsection{Bayesian Game} \label{sec:bayesiangame}
We begin with the formal definition of a Bayesian game, followed by the introduction of its solution concept, the Bayesian Nash equilibrium. Finally, we present an example that extends the previous examples.

Although sometimes introduced in terms of the nature states \citep{Osborne1994}, we adopt the Bayesian game in terms of players' types as originally proposed \citep{harsanyi1967games,fudenberg1991game}.

\begin{definition}
A \textit{Bayesian game} is a tuple $\Gamma = \{M, \bm{A}, T, P, \bm{U}\}$, such that:
\begin{itemize}
    \item $M = \{1, \ldots, m\}$ is the finite player set.
    \item $\bm{A}$ represents the set of action profiles, where $A^i$ is the action set of player $i$ for $i \in M$.
    \item $T^i$ is the finite type set of player $i$, and $t^i \in T^i$ its type. $T=(T^1,\dots,T^m) $ is called the type profile tuple of $\Gamma$.
    \item $P: T \rightarrow [0, 1]$ is a probability distribution over $T$, referred to as the common prior. The belief of player $i$ is denoted by
    \[
    P(t^{-i} \mid t^i) = \frac{P(t^{-i}, t^i)}{P(t^i)} = \frac{P(t^{-i}, t^i)}{\sum_{t^{-i}} P(t^{-i}, t^i)},
    \]
    which describes player $i$’s uncertainty about the other $m - 1$ players’ possible types $t^{-i}$, given player $i$’s type $t^i$, where $t^{-i} = (t^1, \ldots, t^{i-1}, t^{i+1}, \ldots, t^m)$ represents the tuple of the types of all the players except for player $i$.
    \item \(\bm{U} = \{u^1, u^2, \dots, u^m\}\) is a set of utility functions, where  $u^i:T \times \bm{A} \rightarrow \mathbb{R}$ is the payoff function, which maps each action profile $\bm{a} \in \bm{A}$ to the pay-off of player $i$ under each type profile $t^i \in T^i$.
\end{itemize}
\end{definition}

Now that the structure of the game is contingent on the types of the players, the concept of a behavioral strategy can naturally be extended to account for these types: $\sigma^i(t^i) :=\sigma^i(A^i\mid t^i)$\citep{bajoori2016behavioral}. This allows the definition of a \emph{Bayesian Nash equilibrium}. 

\begin{table}[!b]
\centering
\caption{Utilities in Deterring Game in Bayesian Game.}
\begin{tabular}{cc|c|c| c |c|c|}
   & \multicolumn{1}{c}{} & \multicolumn{2}{c}{Adversary Type $t^1$: $p=\frac{1}{4}$} & \multicolumn{1}{c}{} & \multicolumn{2}{c}{Adversary Type $t^2$: $p=\frac{3}{4}$} \\
   & \multicolumn{1}{c}{} & \multicolumn{1}{c}{$a$} & \multicolumn{1}{c}{$\neg a$} & \multicolumn{1}{c}{} &\multicolumn{1}{c}{$a$} & \multicolumn{1}{c}{$\neg a$} \\
\cline{3-4} \cline{6-7}
   \multirow{2}*{Deterrer} & $d$ & $(-1000, -1000)$ & $(-1, 1)$& & $(-1, 0)$ & $(-1, 1)$ \\
\cline{3-4} \cline{6-7}
   & $\neg d$ & $(1, -1)$ & $(0, 0)$ && $(1, -1)$ & $(0, 0)$ \\
\cline{3-4} \cline{6-7}
\end{tabular}
\label{tab:nato_bayes}
\end{table}

\begin{definition}
A strategy profile \(\hat{\sigma} = (\hat{\sigma}^1, \ldots, \hat{\sigma}^n)\) is a \textit{Bayesian Nash equilibrium} if for every player $i\in M$ and type $t^i \in T^i$:
\begin{align*}
    \hat{\sigma}^i(t^i)\in \argmax_{{\sigma}^i\in \bm{\Sigma}^i}P(t^{-i}\mid t^i)u^i(t^i,t^{-i},\sigma^i,\hat{\sigma}^{-i}(t^{-i})).
\end{align*}
\end{definition}

\begin{example}[Deterring Game in Bayesian Form] \label{ex:bayesian-form}
    Similar to the normal-form game, we model the actions of the deterring agent $A^1$ and its adversary $A^2$ independently. This time, the adversary can assume different types, where it is either protected against retaliation $(t^2)$ or not $(t^1)$. The relations of the game are illustrated by Figure \ref{fig:bayesian-form} and the game matrices displayed in Table \ref{tab:nato_bayes}. The Bayesian Nash equilibrium is (($\neg a$,$\neg a$), $\neg d$).     
\end{example}

\subsection{Practical Guide of Game Theory} \label{subsec:pracgt}
Game theory offers a structured framework for organizing information on actors' decision-making processes \citep{hermans2014usefulness}. Before applying game-theoretic concepts, a qualitative process should define the specific rules of the game for a given problem. This involves identifying stakeholders, outlining potential policy options, and establishing their interdependencies. Such an approach to framing policy problems is known as metagame analysis \citep{howard1971paradoxes}.

\begin{figure}[!b]
\centering
\tikzset{methods/.style={draw, fill=ibmlight23, text width=7em, text centered, minimum height=3em, rounded corners}}
\def\blockdist{2.3}
\def\edgedist{2.5}
\def\blockdist{2.3}
\def\edgedist{2.5}
\begin{tikzpicture}[->,>=stealth',auto,node distance=3cm,
  thick,main node/.style={circle,draw,font=\sffamily\Large\bfseries}]

    \node (osa) [componprop,text width=8em] {\textbf{Extensive-Form Game}};
    \path (osa.north)+(3.85,0) node (ass1) {\textbf{Game Theory Models}}; 
    \path (osa.south)+(0,1.2) node (seq) [property] {\textbf{Sequential}}; 
    \path (osa.south)+(0,0.5) node (cha) [property] {\textbf{Imperfect Info}};
    \path (osa.east)+(2.2,0) node (nmf) [componprop2,text width=8em ] {\textbf{Normal-Form Game}};
    \path (osa.east)+(2.2,-0.4) node (sim1) [property] {\textbf{Simultaneous}};
    \path (nmf.east)+(2.2,0) node (bg) [componprop, text width=8em] {\textbf{Bayesian\\Game}};
    \path (bg.north)+(0,-1.55) node (sim2) [property] {\textbf{Simultaneous}};
    \path (bg.north)+(0,-2.2) node (priv) [property] {\textbf{Incomplete Info}};

    \path (osa.north)+(4,-5.5) node (elic) {\textbf{Elicitation Requirements}};  
    \path (osa.north)+(0,-4.5) node (probs) [methods,text width=7.5em] {\textbf{Information Sets}};
    \path (nmf.north)+(0,-4.1) node (util) [methods,text width=7.5em ] {\textbf{Utilities}};
    \path (bg.north)+(0,-4.5) node (types) [methods, text width=7.5em] {\textbf{Types and Priors}};

    \path[every node/.style={font=\sffamily\small}]
        (osa.south east) edge[bend left=0] node [right] {} (util.north);    
    \path[every node/.style={font=\sffamily\small}]
        (cha.south) edge[bend left=0] node [right] {} (probs.north);
    \path[every node/.style={font=\sffamily\small}]
        (nmf.south) edge[bend left=0] node [right] {} (util.north);
    \path[every node/.style={font=\sffamily\small}]
        (bg.south west) edge[bend right=0] node [right] {} (util.north);
    \path[every node/.style={font=\sffamily\small}]
        (priv.south) edge[bend left=0] node [right] {} (types.north);
    \begin{pgfonlayer}{background}
        \path (osa.north west)+(-0.3,0.4) node (a) {};
        \path (osa.south east)+(+8,-0.2) node (b) {};
        \path[fill=ibmlight54, draw=ibmlight54, dashed]
            (a) rectangle (b);
            
        \path (elic.north west)+(-3.5,1.5) node (a) {};
        \path (elic.south east)+(+3.1,-0) node (b) {};
        \path[fill=ibmlight2, draw=ibmlight22, dashed]
            (a) rectangle (b);
    \end{pgfonlayer}
\end{tikzpicture}
\caption{Elicitation requirements for different game theory models: while the yellow blocks correspond to the different games along with their associated properties, the purple blocks indicate the different pieces of information that are required to be elicited. The arrows indicate what information is relevant to each game.}
\label{fig:gtprac}
\end{figure}
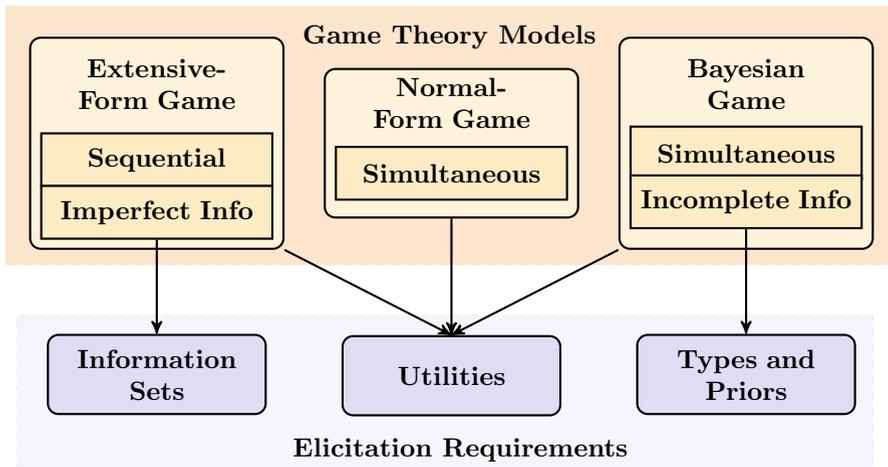

Practitioners should select the game type that best fits the policy problem's structure. Simultaneous decisions suit normal-form and Bayesian games\footnote{Although atypical, Bayesian games have been extended to sequential decision-making \citep{Osborne1994}.}, whereas sequential decisions align with extensive-form games. Additionally, extensive-form games may involve imperfect information, while Bayesian games feature incomplete information. Although selecting a specific game type is useful, some scholars argue that a robust analytical approach benefits from modeling pluralism rather than strict uniformity \citep{allan1999international}, advocating for the use of diverse game-theoretic frameworks where necessary while ensuring that variation remains focused on essential structural distinctions.


As the mathematical dissection of different games shows, each game requires specific information to be gathered before implementation. All games need the elicitation of utilities or preferences, with extensive-form and Bayesian games requiring more detailed utility structures. While utility elicitation is a demanding task, it has been well-studied and we refer the reader to \cite{wakker1996eliciting} for details about elicitation methods. If preference relations are uncertain, alternative methods can still provide insights into the stability of certain solution concepts \citep{li2004preference}.

In extensive-form games, information sets—clarifying who knows what at each decision point—need to be extracted. Additionally, chance nodes require probability distributions for the uncertainties they introduce. For Bayesian games, prior and posterior probabilities of types must also be elicited, a challenging process with guidelines written by \cite{Klami2023}. The characteristics of the different games, along with the required elicitation information can be observed in Figure \ref{fig:gtprac}.

To bridge the gap between theory and application in game theory, a growing body of work focuses on deriving game-theoretic models from simulation data to enhance empirical validity. A comprehensive survey of recent advances in this area is provided by \cite{wellman2024empirical}.

\section{Causality} \label{sec:caus}

The concept of causality pertains to the underlying regularities within a given environment (or context) that go beyond mere probabilistic (or associative) relationships. A causal relationship enables assessing how a change in the cause affects the outcome, thereby establishing a directional order between causally linked variables \citep{Spirtes2000,tesisgonzalezsoto}. Attempts to understand the meaning of cause have troubled philosophers and scientists,
from David Hume and John Stuart Mill to modern scientists as Suppes \citep{suppes1973probabilistic}, Cartwright \citep{cartwright1983laws}, Spirtes \citep{Spirtes2000}, Pearl \citep{Pearl2009a}, Spohn \citep{spohn2012laws}. We adopt the manipulationist interpretation of causality \citep{Spirtes2000,woodward2005making}, as this framework is also adopted in studies of the intersection between causality and game theory \citep{Soto2020, hammond2023reasoning}. The main paradigm of this interpretation has been expressed by \citeauthor{cook1979quasi} as: manipulation of a cause will result in a corresponding change in the effect. Similar descriptions of the manipulationist approach can be found in the work of \citeauthor{holland1986statistics}. 

\begin{figure}[!b]
\centering
\begin{tikzpicture}[->,>=stealth',auto,node distance=3cm,
  thick,main node/.style={circle,draw,font=\sffamily\Large\bfseries}]

  \path \compon{1}{\textbf{Observational Data}};
  \path (p1.south)+(0,-0.25) node (empty1) {}; 
  
  \path (p1)+(-4.,-0.8) \assum{2}{\textbf{Causal Discovery Assumptions}};
  \path (p2.north)+(0,0.45) node (text) [text width=4cm, align=center] {\textbf{Assumptions}}; 
  \path (p1)+(4,-0.8) \task{3}{\textbf{Domain Knowledge}};
  \path (p3.north)+(0,0.45) node (text) [text width=4cm, align=center] {\textbf{Expert Knowledge}};

  \path (p1.south)+(0.0,-1.5) \compon{4}{\textbf{Causal Bayesian Network}};
  \path (p4.south)+(0,-0.15) node (empty2) {}; 

  \path (p4)+(-4.0,-0.8) \assum{5}{\textbf{Causal Inference Assumptions}};
  \path (p4)+(4,-0.8) \task{6}{\textbf{Probability Elicitation}};

  \path (p4.south)+(0.0,-1.5) \compon{7}{\textbf{Causal Estimands and Estimates}};
  \path (p7.south)+(0,-0.45) node (text) [text width=4cm, align=center] {\textbf{Causal Objects}}; 

  \path [line] (p1.south) -- coordinate[pos=0.5] (a1) node [above]  {} (p4);
  \path[every node/.style={font=\sffamily\small}]
        (p2.east) edge[bend left=0] node [right] {} (empty1.center); 
  \path[every node/.style={font=\sffamily\small}]
        (p3.west) edge[bend left=0] node [right] {} (empty1.center); 
  \path [line] (p4.south) -- coordinate[pos=0.5] (a2) node [above]  {} (p7);
  \path[every node/.style={font=\sffamily\small}]
        (p5.east) edge[bend left=0] node [right] {} (empty2.center); 
  \path[every node/.style={font=\sffamily\small}]
        (p6.west) edge[bend left=0] node [right] {} (empty2.center); 
  \path[every node/.style={font=\sffamily\small\itshape}, dashed]
        (p1.east) edge[bend left=30] node [right] {Double Dipping} (p7.north east); 

  \begin{scope}[on background layer]
    \path (p2.north west)+(-0.35,0.8) node (a) {};
        \path (p5.south east)+(0.35,-0.2) node (b) {};
        \path[fill=ibmlight1, draw=ibmlight1, dashed]
            (a) rectangle (b);
    \path (p3.north west)+(-0.4,0.8) node (a) {};
        \path (p6.south east)+(0.4,-0.2) node (b) {};
        \path[fill=ibmlight2, draw=ibmlight2, dashed]
            (a) rectangle (b);
    \path (p4.north west)+(-0.35,0.2) node (a) {};
        \path (p7.south east)+(0.35,-0.8) node (b) {};
        \path[fill=ibmlight5, draw=ibmlight5, dashed]
            (a) rectangle (b);
  \end{scope}
\end{tikzpicture}
\caption{Flow of causal reasoning from data to graphical components (causal discovery) and subsequently to causal estimands and estimates (causal inference). While the light blue part indicates at what stage assumptions have to be taken into account, the purple part indicates possible supplementary information from domain experts. Double dipping occurs when data in the causal discovery stage is being reused at the causal inference stage.}
\label{fig:causalprac}
\end{figure}
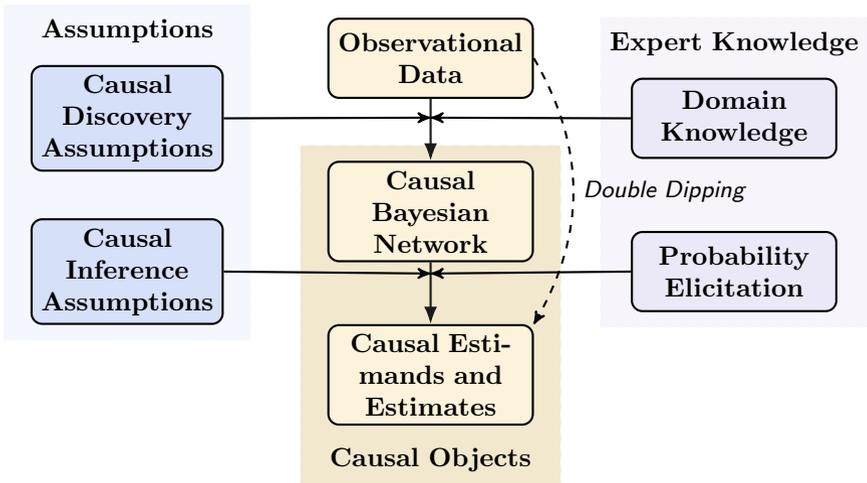

We \textit{restrict} ourselves to the aspects of causality that are most relevant to the specification and practical utility of such models.\footnote{Although some game-theoretical advances have been made at the counterfactual layer of Pearl's Causal Hierarchy \citep{hammond2023reasoning,Bareinboim2020a}, we exclude them here as their practical utility is under scrutiny \citep{Bareinboim2020a}.} Therefore, we first discuss some essential preliminaries. We continue with essential concepts such as the causal Bayesian network, the interventional distribution, and its associated truncated factorization in Section \ref{CBNprelim}. These concepts help to formalize causal structures, define causal interventions, and compute the effects of interventions, respectively. We then elaborate on the process of learning causal graphs, known as causal discovery, in Section \ref{subsec:discovery} and continue with inferring causal effects, known as causal inference, in Section \ref{subsec:praccausal}. The focus of the latter two sections will be on their practical considerations, which are summarized in Figure \ref{fig:causalprac}.

\subsection{Preliminaries} 
A graph is denoted by $G=(\bm{V},\bm{E})$, where $\bm{V}=\{V_1,\dots,V_n\}$ is the set of vertices (or nodes) and $\bm{E}$ the set of edges. A graph can be \textit{directed} when every edge has a direction, \textit{undirected} when no edge has a direction or \textit{partially directed} when some but not all edges have a direction. A graph can contain a \textit{cycle} when there exists a directed path from a node to itself. When there is no such path and the graph is directed, we call this a \textit{directed acyclic graph} (DAG). Edges can also be \textit{bidirected}. A graph containing only directed and bidirected arrows without directed cycles is called an \textit{acyclic directed mixed graph} (ADMG). When there is a directed edge from node $V_1$ to node $V_2$ (or $V_1\xrightarrow[]{}V_2$, in short), we say that $V_1$ is a \emph{parent} of $V_2$ and $V_2$ a \emph{child} of $V_1$. The set of parents of $V_2$ is denoted by $\textbf{\text{pa}}(V_2)$ and the set of children of $V_1$ is denoted by $\textbf{\text{ch}}(V_1)$.

When graphs are endowed with probabilistic meaning, random variables $\bm{X}=\{X_1,\dots,X_n\}$ will correspond to nodes of the graph $\bm{V}=\{V_1,\dots,V_n\}$ and therefore $\bm{V}$ will inherit the probability distributions and state spaces from $\bm{X}$ (meaning $P(\bm{V})$ and $v_i$ will correspond to $P(\bm{X})$ and $x_i$, respectively). In this case, $\textbf{\text{pa}}(V_i)$ refers to the random variables that are associated with the parents of $V_i$. The assignment of random variables $\textbf{\text{pa}}(V_i)$ is denoted by $\bm{pa}_i$. 

\subsection{Bayesian Network and Interventional Distribution} \label{CBNprelim}

A \textit{Bayesian network} (BN) represents a distribution over a set of random variables by associating it to a graph in such a way that each of the random variables corresponds to a node in the graph, and the edges describe the probabilistic dependencies found in the distribution. Formally, a Bayesian network contains a directed acyclic graph $G$ whose nodes represent random variables $X_1 , ... , X_n$ and a probability distribution $P$ that factorizes over $G$, which means that the joint probability distribution of random variables $X_i\in \bm{X}$ can be expressed as
\begin{align*}
P(x_1,\dots x_n) = \prod_{i=1}^n P(x_i\mid \bm{pa}_i).
\end{align*}
This factorization of the distribution implies that every random variable $X_i$ is conditionally independent of its non-descendants given its parents in the graph.

A \emph{causal Bayesian network} is a BN whose edges represent relations of cause-eﬀect. Also, the model is enriched
with an operator named $do()$, which is defined over graphs, and whose action is described as follows.

\begin{definition}
Let $\bm{X_S}$ and $\bm{X_K}$ be sets of random variables. The \emph{interventional distribution} $P(\bm{x_S}\mid do(\bm{X_K}=\bm{x_K}))$ encodes the probability that $\bm{X_S}=\bm{x_S}$ given that $\bm{X_K}$ is forced to take value $\bm{x_K}$ (denoted by the \emph{do-operator} $do(\bm{X_K}=\bm{x_K})$, or $do(\bm{x_K})$ in short) with probability 1.
\end{definition}

The behavior of the $do$-operator within a causal BN leads to a \emph{truncated factorization} of the distribution \citep{Bareinboim2012}:

\begin{definition}
 Let $P(x_1, \dots, x_n)$ be the joint probability distribution of random variables $X_i\in \bm{X} $ corresponding to the nodes $V_i\in \bm{V}$ in the directed acyclic graph $G=(\bm{V},\bm{E})$. Suppose $\bm{X_K}, \bm{X_S}\subset{\bm{X}}$ and $\bm{X_K}\bigcap \bm{X_S}=\emptyset$. Then:

\begin{equation*}
    P(\bm{x_S}\mid do(\bm{X_K} = \bm{x_K}))=\sum_{x_j\mid j\notin \bm{K}\bigcup\bm{S}}\prod_{j\notin \bm{K}} P(x_j\mid \bm{pa}_j).
\end{equation*}   
\end{definition}

Although the introduced definitions pertain to \emph{Markovian} Causal Bayesian networks, they also extend naturally to scenarios involving \emph{semi-Markovian} models where unobserved confounding variables are present. In this case, confounding relations can be represented by bidirected arrows in an ADMG and a similar truncated factorization can be derived \citep{richardson2014factorization}. Next, we demonstrate the application of the truncated factorization by refining our previous example within a causal framework.

\begin{example}[Deterring Relations in Causal Form] \label{ex:causal}
Suppose we have observational data on the explicitness of deterrence messages $X_D$, which can either be explicit ($d$) or vague ($\neg d$).  The goal of an explicit message is to dissuade the adversary from committing to an aggressive operation $X_A$ ($X_D \rightarrow X_A$). However, both the explicitness of the deterrence message and the adversary's decision to attack are shaped by the deterring agent's military and strategic capabilities $X_C$, which can be strong ($c$) or weak ($\neg c$). These capabilities influence the explicitness of the message $X_C\rightarrow X_D$, but also directly affect the adversary's decision to commit to an aggressive operation $X_C \rightarrow X_A$. The causal relations are displayed in Figure \ref{fig:causal_graph}.

To compute the causal effect of explicit deterrence messaging on successful dissuasion, the covariate, the deterrer's capabilities, should be adjusted for. Therefore, the $do$-operator can be deployed, and the truncated factorization can be utilized:

\begin{align*}
P(X_A\mid do(X_D=d))=\sum_{x_C\in \{c,\neg c\}}P(X_A\mid X_D=d, X_C=x_C)P(X_C=x_C).
\end{align*}

\end{example}

\subsection{Practical Guide to Causal Discovery} \label{subsec:discovery}
The goal of causal discovery is to recover the causal structure $G=(\bm{V},\bm{E})$, either from observational data, interventional data, or a mixture of both. While there already exists a plethora of survey papers on causal discovery \citep{zanga2022survey, nogueira2022methods, assaad2022survey}, including an exhaustive list of assumptions \citep{vonk2023disentangling} and a practical guide to causal discovery \citep{malinsky2018causal}, we would like to limit the discussions to the following \textit{three} practical points when bringing causal discovery algorithms into practice.

First, when engaging in causal discovery, practitioners should abide by different assumptions. Causal sufficiency, the assumption that the dataset in question contains all common causes of the variables included, bifurcates different methods of causal discovery. Sometimes, assumptions can be replaced by others. For example, faithfulness, the assumption that dependencies and independencies in the data are solely due to the underlying causal structure, is considered a strong assumption \citep{Andersen2013} and can be replaced by assumptions on the data-generating process of the graph \citep{Shimizu2006}. Therefore, it is recommended to check the validity of the assumptions per application and choose the appropriate causal discovery algorithm accordingly. 

Second, the quality and the size of the available data should be leading when bringing the causal discovery methods into practice. Causal discovery algorithms are mostly grouped into constraint-based methods or score-based methods. While constraint-based methods exploit the conditional independencies to inform the structure of the graph \citep{Spirtes1990, Spirtes1991}, score-based methods iteratively adjust the graph, scoring and comparing different structures to find the best fit to the data \citep{Chickering2002}. As both methods rely on statistical tests, larger sample sizes are typically preferred, because they provide more reliable estimates and increase the power of the tests. Nonetheless, score-based methods are preferred in the case of small sample sizes \citep{malinsky2018causal}. Despite the goal of causal discovery to specify the dependency \emph{between} the random variables, the observations in the available sample are assumed to be independently distributed in order for these causal discovery algorithms to work \citep{vonk2023disentangling}. 

Finally, research has shown that causal discovery algorithms can be unstable \citep{kitson2024impact} or that only limited parts of the causal graph can be discovered from pure observations \citep{Peters2017}. To this purpose, domain knowledge can be used to refine the performance of causal discovery algorithms and has been incorporated via tiered background knowledge \citep{pmlr-v108-andrews20a} user interactions \citep{bg-19-2095-2022} or the penalization of the search process \citep{pmlr-v182-hasan22a}. It is recommended that practitioners assess the possibility of incorporating domain knowledge or experts to enhance the quality of the obtained causal graphs.

\subsection{Practical Guide to Causal Inference} \label{subsec:praccausal}
Causal inference is concerned with the estimation of an outcome variable under possible alternations. Similar to causal discovery, as sufficient survey papers have been dedicated to causal inference \citep{ding2018causal,yao2021survey,Pearl2009b}, we strict ourselves to \textit{three} practical points when engaging in causal inference.

First, identifying the necessary components to address inference queries of interest requires a sound graphical structure. This graphical structure can be obtained via domain experts or causal discovery methods. When causal inference methods are applied to data that has already undergone causal discovery algorithms, this can result in \emph{double dipping}, compromising the validity of the confidence intervals provided by the statistical methods \citep{cheng2024data}. Practitioners should be mindful of this issue and, if needed, apply available methods that can correct for this bias \citep{gradu2024valid}.

Second, it can be the case that there is no data available to estimate the conditional probability distributions (or one does not want to engage in double dipping). In this case, practitioners can still engage in causal inference when the conditional probability distributions are elicited from domain experts. There already exist efficient methods to infer discrete conditional probability distributions from experts \citep{barons2022balancing,alkhairy2020quantifying,hassall2019facilitating}, but eliciting continuous conditional probability distributions remains underdeveloped. 

Finally, causal inference depends significantly on statistical methods, and, as with causal discovery, larger sample sizes generally yield more reliable estimates. Since the causal structure requires adjusting for specific covariates when estimating causal effects, practitioners should be aware that such adjustments can further reduce the effective sample size or result in positivity violations \citep{DAmour2021} in the non-parametric causal inference case. Alternatively, parametric causal inference can be used at the cost of assuming a specific parametric form for the underlying distribution. Therefore, some semi-parametric approaches to causal inference have been developed \citep{bhattacharya2022semiparametric,luo2023semiparametric}.

\section{Game Theory and Causality} \label{sec:int}
In this section, we explore the intersection of game theory and causality by integrating causal concepts from the previous section into the decision-making framework. Specifically, we extend Bayesian networks to influence diagrams \citep{howard2005influence}, distinguishing between purely probabilistic structures and decision-theoretic elements. This framework is further generalized to multi-agent settings, leading to multi-agent influence diagrams \citep{KOLLER2003181, hammond2023reasoning}, which form the foundational structure of causal games. To account for uncertainty in the causal structure of a game, we adapt the approach of \cite{soto2019choosing}, introducing the causal Bayesian game. Finally, we discuss key considerations for the practical implementation of these models.

\subsection{Influence Diagram}
An influence diagram extends a Bayesian network to the decision realm by dissecting the nodes into chance nodes, utility nodes, and decision nodes. More formally:
\begin{definition}
    An \emph{influence diagrams} contains a graphical structure $G=(\bm{V},\bm{E})$ where $\bm{V}$ is separated in decision nodes $\bm{D}$, chance nodes $\bm{X}$ and utility nodes $\bm{U}$. While the conditional probability distributions of $\bm{X}$ and $\bm{U}$ are known, any \emph{decision rule} $\sigma(D)$ with $D\in \bm{D}$ corresponds to a conditional probability distribution over the decisions and hence all decision rules constitute the full joint probability distribution $P(\bm{V})$.
\end{definition}

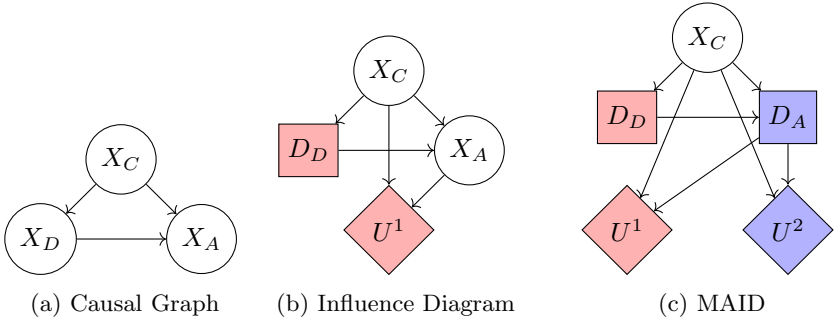
\begin{figure}[!t]
    \centering
    \begin{minipage}[b]{.3\textwidth}
        \centering
        \begin{subfigure}[b]{\textwidth}
            \centering
            \begin{tikzpicture}[bnnode/.style={draw,circle, minimum size=5mm, node distance=1.5cm}]
    \node[bnnode] (d) {$X_D$};
    \node[bnnode,above right of=d] (c) {$X_C$};
    \node[bnnode,below right of=c] (a) {$X_A$};
    \draw[->] (d) to (a);
    \draw[->] (c) to (d);
    \draw[->] (c) to (a);
\end{tikzpicture}
            \subcaption{Causal Graph}
            \label{fig:causal_graph}
        \end{subfigure}\\
    \end{minipage}%
    \begin{minipage}[b]{.3\textwidth}
        \centering
        \begin{subfigure}[b]{\textwidth}
            \centering
            \begin{tikzpicture}[
    bnnode/.style={draw, circle, minimum size=5mm, node distance=1.5cm},
    squarednode/.style={draw, rectangle, fill=red!30, minimum size=7mm, node distance=1.5cm},
    diamondnode/.style={draw, diamond, fill=red!30, minimum size=7mm, node distance=1.5cm},
    circlenode/.style={draw, circle, fill=white, minimum size=7mm, node distance=1.5cm}
]

\node[squarednode] (d) {$D_D$};
\node[diamondnode, below right of=d] (u) {$U^1$};
\node[circlenode, above right of=d] (c) {$X_C$};
\node[circlenode, below right of=c] (a) {$X_A$};

\draw[->] (d) to (a);
\draw[->] (c) to (d);
\draw[->] (c) to (a);
\draw[->] (c) to (u);
\draw[->] (a) to (u);

\end{tikzpicture}
            \subcaption{Influence Diagram}
            \label{fig:influence}
        \end{subfigure}\\
    \end{minipage}%
    \begin{minipage}[b]{.4\textwidth}
        \centering
        \begin{subfigure}[b]{\textwidth}
            \centering
            \begin{tikzpicture}[
    redbox/.style={draw, rectangle, fill=red!30, minimum size=7mm, node distance=1.5cm},
    bluebox/.style={draw, rectangle, fill=blue!30, minimum size=7mm, node distance=1.5cm},
    reddiamond/.style={draw, diamond, fill=red!30, minimum size=7mm, node distance=1.5cm},
    bluediamond/.style={draw, diamond, fill=blue!30, minimum size=7mm, node distance=1.5cm},
    circlenode/.style={draw, circle, fill=white, minimum size=7mm, node distance=1.5cm}
]

\node[redbox] (d) {$D_D$};            
\node[reddiamond, below of=d] (ud) {$U^1$};  
\node[bluediamond, below of=a] (ua) {$U^2$}; 
\node[circlenode, above right of=d] (c) {$X_C$}; 
\node[bluebox,below right of=c] (a) {$D_A$};      

\draw[->] (d) to (a);
\draw[->] (c) to (d);
\draw[->] (c) to (a);
\draw[->] (c) to (ua);
\draw[->] (c) to (ud);
\draw[->] (a) to (ud);
\draw[->] (a) to (ua);

\end{tikzpicture}
            \subcaption{MAID}
            \label{fig:MAID}
        \end{subfigure}\\
    \end{minipage}%
    \caption{The relations of the causal graph (a), influence diagram (b), and multi-agent influence diagram (c) of Example \ref{ex:causal}, \ref{ex:influence} and \ref{ex:MAID}, respectively. The causal graph does not distinguish between chance, decision and utility nodes as do the influence diagrams. In addition, the multi-agent influence diagram models the decision of the adversary also strategically. }
    \label{fig:game_theory-examples}
\end{figure}

\begin{example}[Deterring Relations as Influence Diagram]  \label{ex:influence}
    A refinement of Example \ref{ex:causal} involves separating the chance node of the military and strategic capabilities $X_C$ from the decision node of deterrence messaging $D_C$. In this framework, the aggressor's decision to conduct an aggressive operation $X_A$ can be modeled as another chance node, which is followed by a final utility node of the deterring agent $U^1$. These relations are illustrated by Figure \ref{fig:influence}.
\end{example}

Influence diagrams incorporate decision-making elements into Bayesian networks, but they only model the decision-making of a single agent, meaning there is no strategic interaction between agents. Strategic considerations emerge only when multiple agents make decisions in response to each other, as seen in multi-agent influence diagrams.

\subsection{Multi-Agent Influence Diagram And Causal Game}
We start by introducing multi-agent influence diagrams \citep{KOLLER2003181} and continue with the definition of a causal game \citep{hammond2023reasoning}. The latter concept will then be illustrated with an example.

\begin{definition}
    A \emph{multi-agent influence diagram} (MAID) contains a graphical structure $G=(\bm{V},\bm{E})$ and a set of agents $M=\{1,\dots,m\}$. Furthermore, the nodes $\bm{V}$ are separated in decision nodes $\bm{D}=\cup_{i\in M}\bm{D}^{i}$, chance nodes $\bm{X}$ and utility nodes $\bm{U}=\cup_{i\in M}\bm{U}^{i}$. Each strategy $\sigma^i(D^i)$ with $D^i\in \bm{D^i}$ defines a conditional probability distribution over a decision node. Consequently, given conditional probability distributions of $\bm{X}$ and $\bm{U}$, a complete strategy profile $\sigma$ constitutes the full joint probability distribution $P(\bm{V})$.\footnote{Since EFGs and MAIDs are proven to be equivalent \citep{hammond2023reasoning}, we use $\sigma$ again for a strategy profile.}
\end{definition}

The causal game $\Gamma$ associated with a MAID can be seen as a more abstract form of a MAID, where the parameters of the decision variables are yet to be defined. 

\begin{definition}
    A \emph{causal game} $\Gamma$ is a MAID such that for any chosen strategy profile $\sigma$, the induced joint probability distribution $P^\sigma(\bm{V})$ corresponds to a causal Bayesian network.
\end{definition}

Similarly to extensive-form games, the Nash equilibrium of causal games can be defined in terms of the strategy profiles:

\begin{definition}
A strategy profile \(\hat{\sigma} = (\hat{\sigma}^1, \ldots, \hat{\sigma}^n)\) is a \textit{Nash equilibrium} if for every player $i\in \{1,\ldots,m \}$:
\begin{align*}
    \hat{\sigma}^i\in \argmax_{{\sigma}^i\in \bm{\Sigma}^i}\sum_{U\in \bm{U}^i}\mathbb{E}_{[\sigma^i,\hat{\sigma}^{-i}]}[U].
\end{align*}
\end{definition}

Multi-agent influence diagrams are powerful models, because they allow for the calculation of equilibria as well as the computation of policy interventions.

\begin{figure}[!b]
    \centering
    \caption{The utilities of the deterring agent (left) and attacking agent (right).}
    \begin{minipage}[b]{.5\textwidth}
        \centering
        \begin{tabular}{c||c|c}
                    $U^1$ & $X_C=c$ & $X_C=\neg c$ \\\hline
                    $D_A=a$ & 0 & -1  \\
                    $D_A=\neg a$  & 1 & 1 
        \end{tabular}
    \end{minipage}%
    \begin{minipage}[b]{.5\textwidth}
        \centering
        \begin{tabular}{c||c|c}
                    $U^2$ & $X_C=c$ & $X_C=\neg c$ \\\hline
                    $D_A=a$ & -1000 & 1  \\
                    $D_A=\neg a$  & -1 & -1 
        \end{tabular}
    \end{minipage}%

    \label{fig:utilities_detgame}
\end{figure} 

\begin{example}[Multi-Agent Influence Diagram Deterring Game]   \label{ex:MAID}
    A multi-agent influence diagram can be derived from Example \ref{ex:influence} when the adversary's decision to conduct an aggressive operation $a$ or refrain from one $\neg a$ is modeled as a decision node of another agent $D_A$. This decision node is influenced by the deterring agent's decision node, \(D_D\), which represents whether the deterrence messages are explicit (\(d\)) or vague (\(\neg d\)). Alongside the deterring agent's utility node $U^1$, the adversary also possesses a utility node $U^2$. Both utility nodes depend on the attacker's decision $D_A$ and the deterrer’s capabilities $X_C$, which can be either strong ($c$) or weak ($\neg c$), with an equal probability distribution. The relations are displayed in Figure \ref{fig:MAID} and the utilities are further specified in Table \ref{fig:utilities_detgame}.

    The game has eight Nash equilibria, one of which is equilibrium $\hat{\sigma}$ where the deterring agent issues an explicit deterrence message when possessing strong capabilities and a vague one otherwise. In this scenario, the adversary chooses not to attack if and only if the deterring agent demonstrates strong capabilities regardless of the deterrence message. The equilibrium $\hat{\sigma}$ also happens to be a subgame perfect equilibrium.\footnote{Although the subgame perfect equilibrium of Definition \ref{def:spe} naturally extends to causal games, the notion of subgames in causal games is much richer than in EFGs. We refer the reader to the work of \cite{hammond2023reasoning} for details on this.} Within this equilibrium,  the expected utility of the deterring agent for sending out an explicit message is $\mathbb{E}_{[\hat{\sigma}]}[U^1 \mid D_D = d] = 1$. We can also assess intervention effects in this equilibrium; for instance, if allied agents force an explicit deterrence message regardless of its capability, the utility becomes $\mathbb{E}_{[\hat{\sigma}]}[U^1 \mid do(D_D = d)] = 0$.
\end{example}

This example is a \emph{post-policy} intervention as the results are computed after a strategy profile from the Nash equilibrium has been chosen. In contrast, \emph{pre-policy} interventions allow agents to adjust their strategy profile after an intervention, which requires a more refined notion of a MAID \citep{hammond2023reasoning}. The introduction of these notions, while relevant for strategic reasoning, is considered beyond the scope of the paper as they do not bring additional implications for their practical implementation.

\subsection{Causal Bayesian Games}
The notion of a causal Bayesian game \citep{soto2019choosing} was developed in order to allow for uncertainty about a graphical structure controlling an environment in which agents are located. We refine the definition and notation of \citeauthor{soto2019choosing} to align with the previously introduced causal games while ensuring consistency with earlier introduced Bayesian games. Following the Bayesian game in Section \ref{sec:bayesiangame}, the types correspond to distinct MAIDs within the family of causal graphical structures $\mathcal{G}$. Moreover, as in Bayesian games, players act independently, implying that we consider a subset of MAIDs where no direct causal paths exist between the decision nodes of different agents. Unlike the earlier introduced Bayesian game, players agree on the common state space $\mathcal{G}$ of possible graphical models but have private beliefs about the probability of these states $\mu_i(\mathcal{G})$. Naturally, interventions on decision nodes induce variations in payoffs across different graphical models $G\in \mathcal{G}$.

\begin{definition}
    Consider a family of different causal structures $G\in \mathcal{G}$ where no direct paths exist between the decision nodes of different agents. Each agent $i\in\{1,\dots,m\}$ has a private belief about the probability of these causal structures $\mu_i(\mathcal{G})$ and a higher-order belief $\mu_i(\mu_{-i}(\mathcal{G}))$, which reflect uncertainty over other $m-1$ players' beliefs. Each strategy $\sigma^i(G) = \sigma^i(D^i\mid G)$ defines a conditional probability distribution over a decision node $D^i\in \bm{D^i}$ conditional on the belief $\mu_i(G)$ of that graphical model $G \in \mathcal{G}$ for agent $i\in\{1,\dots,m\}$.\footnote{Although the $do$-operator is applied in the original paper, it is relaxed here to allow for the implementation of more dynamic strategies.} A \emph{causal Bayesian game} $\Gamma$ is a MAID such that for any chosen strategy profile $\sigma$, the induced joint probability distribution $P^\sigma(\bm{V})$ corresponds to a causal Bayesian network.
\end{definition}

Note that the causal Bayesian network is not only induced by the strategies for each player but by the strategies of the players conditioned on the same graphical model. Naturally, these considerations are also reflected in the Bayesian Nash equilibrium.

\begin{definition}
A strategy profile \(\hat{\sigma} = (\hat{\sigma}^1, \ldots, \hat{\sigma}^n)\) is a \textit{Bayesian Nash equilibrium} if for every player $i\in \{1,\ldots,m \}$ and graphical structure $G\in \mathcal{G}$:
\begin{align*}
    \hat{\sigma}^i(G)\in \argmax_{{\sigma}^i\in \bm{\Sigma}^i}\mu_i(\mu_{-i}(G))\sum_{U\in \bm{U}^i}\mathbb{E}_{[\sigma^i(G),\hat{\sigma}^{-i}(G)]}[U].
\end{align*}
\end{definition}

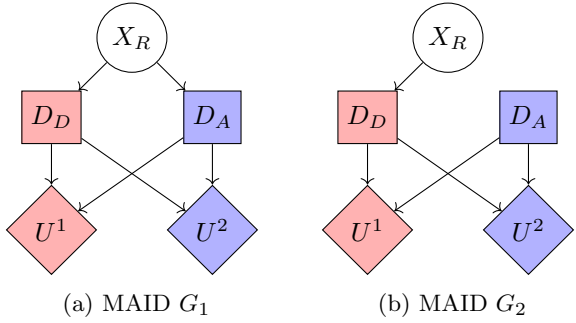
\begin{figure}[!t]
    \centering
    \begin{minipage}[b]{.3\textwidth}
        \centering
        \begin{subfigure}[b]{\textwidth}
            \centering
            \begin{tikzpicture}[
    redbox/.style={draw, rectangle, fill=red!30, minimum size=7mm, node distance=1.5cm},
    bluebox/.style={draw, rectangle, fill=blue!30, minimum size=7mm, node distance=1.5cm},
    reddiamond/.style={draw, diamond, fill=red!30, minimum size=7mm, node distance=1.5cm},
    bluediamond/.style={draw, diamond, fill=blue!30, minimum size=7mm, node distance=1.5cm},
    circlenode/.style={draw, circle, fill=white, minimum size=7mm, node distance=1.5cm}
]

\node[redbox] (d) {$D_D$};            
\node[reddiamond, below of=d] (ud) {$U^1$};  
\node[circlenode, above right of=d] (c) {$X_R$}; 
\node[bluebox, below right of=c] (a) {$D_A$};      
\node[bluediamond, below of=a] (ua) {$U^2$}; 

\draw[->] (c) to (d);
\draw[->] (c) to (a);
\draw[->] (d) to (ud);
\draw[->] (d) to (ua);
\draw[->] (a) to (ud);
\draw[->] (a) to (ua);

\end{tikzpicture}
            \subcaption{MAID $G_1$}
            \label{fig:cba1}
        \end{subfigure}\\
    \end{minipage}%
    \begin{minipage}[b]{.4\textwidth}
        \centering
        \begin{subfigure}[b]{\textwidth}
            \centering
            \begin{tikzpicture}[
    redbox/.style={draw, rectangle, fill=red!30, minimum size=7mm, node distance=1.5cm},
    bluebox/.style={draw, rectangle, fill=blue!30, minimum size=7mm, node distance=1.5cm},
    reddiamond/.style={draw, diamond, fill=red!30, minimum size=7mm, node distance=1.5cm},
    bluediamond/.style={draw, diamond, fill=blue!30, minimum size=7mm, node distance=1.5cm},
    circlenode/.style={draw, circle, fill=white, minimum size=7mm, node distance=1.5cm}
]

\node[redbox] (d) {$D_D$};            
\node[reddiamond, below of=d] (ud) {$U^1$};  
\node[circlenode, above right of=d] (c) {$X_R$}; 
\node[bluebox, below right of=c] (a) {$D_A$};      
\node[bluediamond, below of=a] (ua) {$U^2$}; 

\draw[->] (c) to (d);
\draw[->] (d) to (ud);
\draw[->] (d) to (ua);
\draw[->] (a) to (ud);
\draw[->] (a) to (ua);

\end{tikzpicture}
            \subcaption{MAID $G_2$}
            \label{fig:cba2}
        \end{subfigure}\\
    \end{minipage}%
    \caption{The different causal structures $G_1$ and $G_2$ of the causal Bayesian game as in Example \ref{ex:cbg1}.}
    \label{fig:game_theory-examples}
\end{figure}

\begin{figure}[!t]
    \centering
    \caption{The utilities of the deterring agent (left) and attacking agent (right) for causal structure $G_1$ (top) and $G_2$ (bottom).}
    
    \begin{minipage}[b]{0.45\textwidth}
        \centering
        \begin{tabular}{c||c|c}
            $U^1(G_1)$ & $D_A=a$ & $D_A=\neg a$ \\\hline
            $D_D=d$ & -1000 & 1 \\
            $D_D=\neg d$ & -1 & 0 
        \end{tabular}
    \end{minipage}
    \hspace{0.05\textwidth}
    \begin{minipage}[b]{0.45\textwidth}
        \centering
        \begin{tabular}{c||c|c}
            $U^2(G_1)$ & $D_A=a$ & $D_A=\neg a$ \\\hline
            $D_D=d$ & -1000 & -1 \\
            $D_D=\neg d$ & 1 & 0 
        \end{tabular}
    \end{minipage}
    
    \vspace{1em} 
    
    \begin{minipage}[b]{0.45\textwidth}
        \centering
        \begin{tabular}{c||c|c}
            $U^1(G_2)$ & $D_A=a$ & $D_A=\neg a$ \\\hline
            $D_D=d$ & -1 & -1 \\
            $D_D=\neg d$ & 1 & 0 
        \end{tabular}
    \end{minipage}
    \hspace{0.05\textwidth}
    \begin{minipage}[b]{0.45\textwidth}
        \centering
        \begin{tabular}{c||c|c}
            $U^2(G_2)$ & $D_A=a$ & $D_A=\neg a$ \\\hline
            $D_D=d$ & 0 & 1 \\
            $D_D=\neg d$ & -1 & 0 
        \end{tabular}
    \end{minipage}
    \label{bayesiangametable}
\end{figure}

\begin{example}[Causal Bayesian Deterring Game] \label{ex:cbg1}
    Suppose two agents have their private beliefs about the causal structure of the game they are playing. In both games, players make decisions to defend $D_D$ and attack $D_A$ independently. While the utility of both agents is the result of both agents' actions, the attacking agent's decision can also be shaped by the defending agent's capability to retaliate $X_R$. Figure \ref{fig:game_theory-examples} illustrates the two different causal structures under consideration. While the attacking agent does not have access to the defending agent's retaliation capacity nor does he think the defending agent thinks he has ($\mu_A(G_2)=1$ and $\mu_A(\mu_{D}(G_1))=1$), the defending agent considers a scenario where the attacking has access to his retaliation capacity with equal probability: $\mu_D(G_1)=\mu_D(G_2)=\mu_D(\mu_{A}(G_1))=\mu_D(\mu_{A}(G_2))=\frac{1}{2}$. Taking into account the utilities for the different structures indicated by Table \ref{bayesiangametable}, a Bayesian Nash equilibrium is given by (($\neg a$,$\neg a$), $\neg d$). 
\end{example}

While this game and associated Bayesian Nash equilibrium is similar to Example \ref{ex:bayesian-form}, we emphasize that uncertainty about the causal structure in this example only gives rise to different pay-offs. When alternative causal structures yield distinct payoff configurations and more sophisticated higher-order beliefs are involved, significant complications may arise.

\subsection{Practical Guide to Causal Game Theory}
Analogous to selecting a game-theoretic model, the choice of a causal game-theoretic model should consider whether agents possess private information regarding the causal structure. As the mathematical dissection discerns different types of nodes within causal games, practitioners must also clearly differentiate between decisions, chance events, and utilities. This distinction can be subtle, as illustrated in Example \ref{ex:MAID}: a deterrer's capability, often modeled as a chance node, may not truly qualify as such if the agent has the option to enhance their capabilities. Consequently, this classification requires careful consideration, thoughtfully aligned with the specific research question at hand.

Unlike the standard causal graphs in Example \ref{ex:causal}, which consist solely of chance nodes, causal games with decision nodes do not require the elicitation of conditional probability distributions for those decisions, as they are being solved in response to the adversary. This game-theoretic aspect in causal games thus reduces some of the elicitation burden. The remaining conditional distribution of the chance nodes and the specifications in the utility nodes can be extracted via the elicitation methods introduced in Section \ref{subsec:praccausal} and \ref{subsec:pracgt}, respectively. 

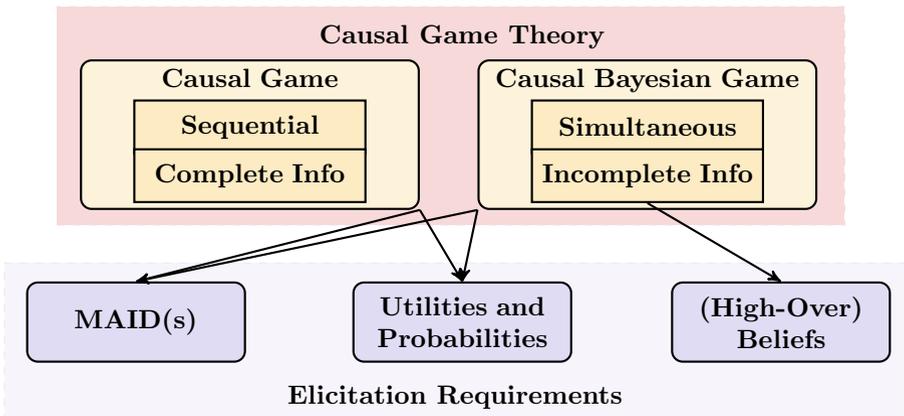
\begin{figure}[!b]
\centering
\tikzset{methods/.style={draw, fill=ibmlight23, text width=7em, text centered, minimum height=3em, rounded corners}}
\def\blockdist{2.3}
\def\edgedist{2.5}
\def\blockdist{2.3}
\def\edgedist{2.5}
\begin{tikzpicture}[->,>=stealth',auto,node distance=3cm,
  thick,main node/.style={circle,draw,font=\sffamily\Large\bfseries}]

    \node (osa) [componprop2, text width=12em] {\textbf{Causal Game}};
    \path (osa.north)+(0,-0.9) node (sim2) [property] {\textbf{Sequential}};
    \path (osa.north)+(0,-1.55) node (sim2) [property] {\textbf{Complete Info}};
    \path (osa.north)+(2.8,0.3) node (ass1) {\textbf{Causal Game Theory}}; 
    \path (osa.east)+(3,0) node (bg) [componprop2, text width=12em] {\textbf{Causal Bayesian Game}};
    \path (bg.north)+(0,-0.9) node (sim2) [property] {\textbf{Simultaneous}};
    \path (bg.north)+(0,-1.55) node (inc) [property] {\textbf{Incomplete Info}};

    \path (osa.north)+(2.7,-4.5) node (elic) {\textbf{Elicitation Requirements}};  
    \path (osa.north)+(-1.5,-3.5) node (probs) [methods,text width=7.5em] {\textbf{MAID(s)}};
    \path (osa.north)+(2.8,-3.5) node (util) [methods,text width=7.5em ] {\textbf{Utilities and Probabilities}};
    \path (osa.north)+(7,-3.5) node (types) [methods, text width=7.5em] {\textbf{(High-Over) Beliefs}};

    \path[every node/.style={font=\sffamily\small}]
        (osa.south east) edge[bend left=0] node [right] {} (util.north);    
    \path[every node/.style={font=\sffamily\small}]
        (osa.south east) edge[bend left=0] node [right] {} (probs.north);
    \path[every node/.style={font=\sffamily\small}]
        (bg.south west) edge[bend right=0] node [right] {} (util.north);
    \path[every node/.style={font=\sffamily\small}]
        (bg.south west) edge[bend right=0] node [right] {} (probs.north);
    \path[every node/.style={font=\sffamily\small}]
        (inc.south) edge[bend left=0] node [right] {} (types.north);
    \begin{pgfonlayer}{background}
        \path (osa.north west)+(-0.3,0.7) node (a) {};
        \path (osa.south east)+(+5.6,-0.2) node (b) {};
        \path[fill=ibmlight55, draw=ibmlight55, dashed]
            (a) rectangle (b);
            
        \path (elic.north west)+(-3.6,1.5) node (a) {};
        \path (elic.south east)+(+3.6,-0) node (b) {};
        \path[fill=ibmlight2, draw=ibmlight22, dashed]
            (a) rectangle (b);
    \end{pgfonlayer}
\end{tikzpicture}
\caption{Elicitation requirements for causal games and causal Bayesian games and associated characteristics: a causal game necessitates the elicitation of a MAID, utilities, and conditional probabilities. In contrast, a causal Bayesian game further requires the elicitation of multiple MAIDs along with beliefs about the probabilities over these graphs and higher-order beliefs about other players' beliefs.}
\label{fig:causalgames}
\end{figure}

While extracting higher-order beliefs in addition to uncertainty over the nature of graphical structure may appear highly complex, the methods for eliciting prior and posterior probabilities outlined in Section \ref{subsec:pracgt} remain applicable \citep{Klami2023}. However, the increased complexity associated with calculating the relevant solution concepts across different causal structures may impede practical implementation.

A summary of the causal game and the causal Bayesian game along with required information for implementation is given in Figure \ref{fig:causalgames}.

\section{Future Directions} \label{sec:conc}
In this paper, we have detailed models from causality, game theory, and their intersection in the context of probabilistic graphical models. In doing so, we have articulated the difference between various models and the required input for their implementation. We have also highlighted research on methods for obtaining this input. Lastly, we have offered detailed examples, practical guidelines, and considerations for their effective implementation.

We propose the following research directions. First, while tools exist for implementing the introduced models and conducting their respective elicitation separately \citep{james_fox-proc-scipy-2021, barons2022balancing}, we recommend developing an integrated elicitation and modeling tool to further streamline the creation of such applications. Second, although causality has been extended to incorporate temporal dynamics \citep{runge2023causal}, no such dynamic causal models, to the best of our knowledge, have been adapted for game-theoretic contexts. Such an extension, though demanding in terms of elicitation requirements, could better capture complex decision-making dynamics.

We hope that our research paves the way for more applications that harness the combined strengths of strategic and causal reasoning.



\bibliography{sn-bibliography.bib}
\clearpage



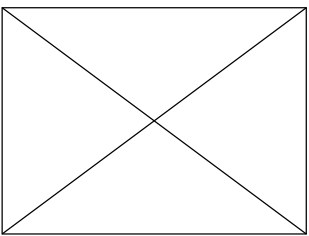
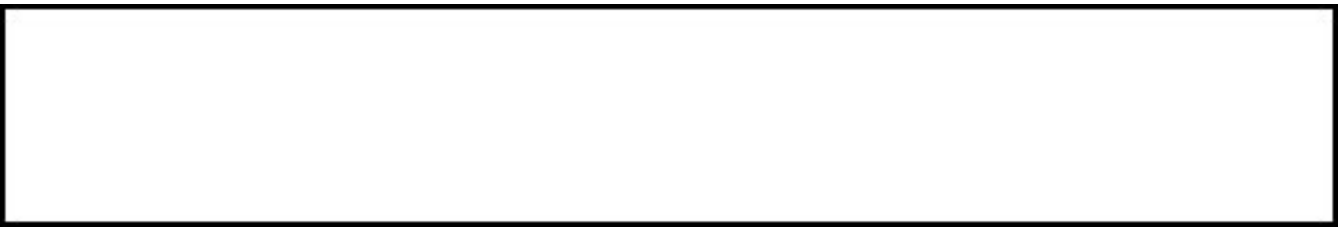
\end{document}